\documentclass[10pt,twocolumn,letterpaper]{article}

\usepackage{cvpr}
\usepackage{times}
\usepackage{epsfig}
\usepackage{graphicx}
\usepackage{amsmath}
\usepackage{amssymb}

\usepackage{times}
\usepackage{epsfig}
\usepackage{float}
\usepackage{url}
\usepackage{subfig}
\usepackage{multirow}
\usepackage{algorithm}
\usepackage{algpseudocode}

\usepackage{array}
\newcolumntype{x}[1]{>{\centering\arraybackslash\hspace{0pt}}m{#1}}
\usepackage{epstopdf}
\usepackage[font=small,labelfont=bf]{caption}

\newcommand{\cc}{\textcolor{black}}
\newcommand{\KG}{\textcolor{black}}

\newcommand{\CR}{\textcolor{black}}
\usepackage[pagebackref=true,breaklinks=true,letterpaper=true,colorlinks=false,bookmarks=false]{hyperref}

\cvprfinalcopy % *** Uncomment this line for the final submission

\def\cvprPaperID{1345} % *** Enter the CVPR Paper ID here

\ifcvprfinal\fi
\begin{document}
	
	%%%%%%%%% TITLE
	\title{FusionSeg: Learning to combine motion and appearance for \KG{fully automatic} segmentation of generic objects in videos}
	%backup title2: How to Segment Thousands of Foreground Object Classes: Generic Object Segmentation in Images }
	
	\author{Suyog Dutt Jain\thanks{Both authors contributed equally to this work} \qquad Bo Xiong\footnotemark[1] \qquad Kristen Grauman \\
		University of Texas at Austin\\
		{\tt\small suyog@cs.utexas.edu, bxiong@cs.utexas.edu, grauman@cs.utexas.edu} \\
		{\tt\small \url{http://vision.cs.utexas.edu/projects/fusionseg/}}
		% For a paper whose authors are all at the same institution,
		% omit the following lines up until the closing ``}''.
		% Additional authors and addresses can be added with ``\and'',
		% just like the second author.
		% To save space, use either the email address or home page, not both
		%\and
		%Kristen Grauman\\
		%Institution2\\
		%First line of institution2 address\\
		%{\small\url{http://www.author.org/~second}}
	}

	%\author{First Author\\
	%Institution1\\
	%Institution1 address\\
	%{\tt\small firstauthor@i1.org}
	%% For a paper whose authors are all at the same institution,
	%% omit the following lines up until the closing ``}''.
	%% Additional authors and addresses can be added with ``\and'',
	%% just like the second author.
	%% To save space, use either the email address or home page, not both
	%\and
	%Second Author\\
	%Institution2\\
	%First line of institution2 address\\
	%{\tt\small secondauthor@i2.org}
	%}
	
	\maketitle

	\begin{abstract}
		We propose an end-to-end learning framework for segmenting generic objects in videos. Our method learns to combine appearance and motion information to produce pixel level segmentation masks for all prominent objects. We formulate the task as a structured prediction problem and design a two-stream fully convolutional neural network which fuses together motion and appearance in a unified framework. Since large-scale video datasets with pixel level segmentations are lacking, we show how to bootstrap weakly annotated videos together with existing image recognition datasets for training. Through experiments on three challenging video segmentation benchmarks, our method substantially improves the state-of-the-art results for segmenting generic \KG{(unseen)} objects. \CR{Code and pre-trained models are available on the project website.}
		%in videos with up to 8% absolute gains in certain cases.
		
		%We propose an end-to-end learning framework for segmenting generic objects in videos. Our method learns to combine appearance and motion information to produce pixel level segmentation masks for all prominent objects in videos. We formulate this task as a structured prediction problem and design a two-stream fully convolutional neural network which fuses together motion and appearance in a unified framework
		%that can be trained in an end-to-end manner. Although lacking large scale video dataset with pixel level segmentations, we propose to utilize large scale image datasets and weakly annotated video dataset.
		%Through experiments on three challenging video segmentation benchmarks we show that our method substantially improves state-of-the-art for segmenting generic objects in videos.

	\end{abstract}
	
	\section{Introduction}\label{sec:introduction}

In video object segmentation, the task is to separate out foreground objects from the background across all frames. This entails computing dense pixel level masks for foreground objects, regardless of the object's category\KG{---i.e., learned object-specific models must \emph{not} be assumed.} A resulting foreground object segment is a spatio-temporal tube delineating object boundaries in both space and time. This fundamental  problem has a variety of applications, including high level vision tasks such as activity and object recognition, as well as graphics areas such as post production video editing and rotoscoping.

In recent years, video object segmentation has received significant attention, with great progress on fully automatic algorithms~\cite{grundmann-cvpr2010,xu-eccv2012,galasso-accv2012,keysegments,ma-cvpr2012,shah-cvpr2013,rehg-iccv2013,ferrari-iccv2013,nlc}, propagation methods~\cite{ren-cvpr2007,tsai-bmvc2010,fathi-bmvc2011,sudheendra-eccv2012,suyog-eccv2014,Wen_2015_CVPR}, and interactive methods% that require a human in the loop continuously to correct errors made by the algorithm
~\cite{wang-tog2005,li-tog2005,bai-2009,Nagaraja_2015_ICCV}. We are interested in the fully automated setup, where the system processes the video directly without any human involvement.  Forgoing manual annotations could scale up the processing of video data, yet it remains a very challenging problem. Automatic algorithms not only need to produce accurate space-time boundaries for any generic object but also need to handle challenges like occlusions, shape changes, and camera motion.

\begin{figure}[t]
\centering
\renewcommand{\tabcolsep}{0pt}
  \captionsetup{ font={footnotesize}, skip=2pt}
\includegraphics[width=1\columnwidth]{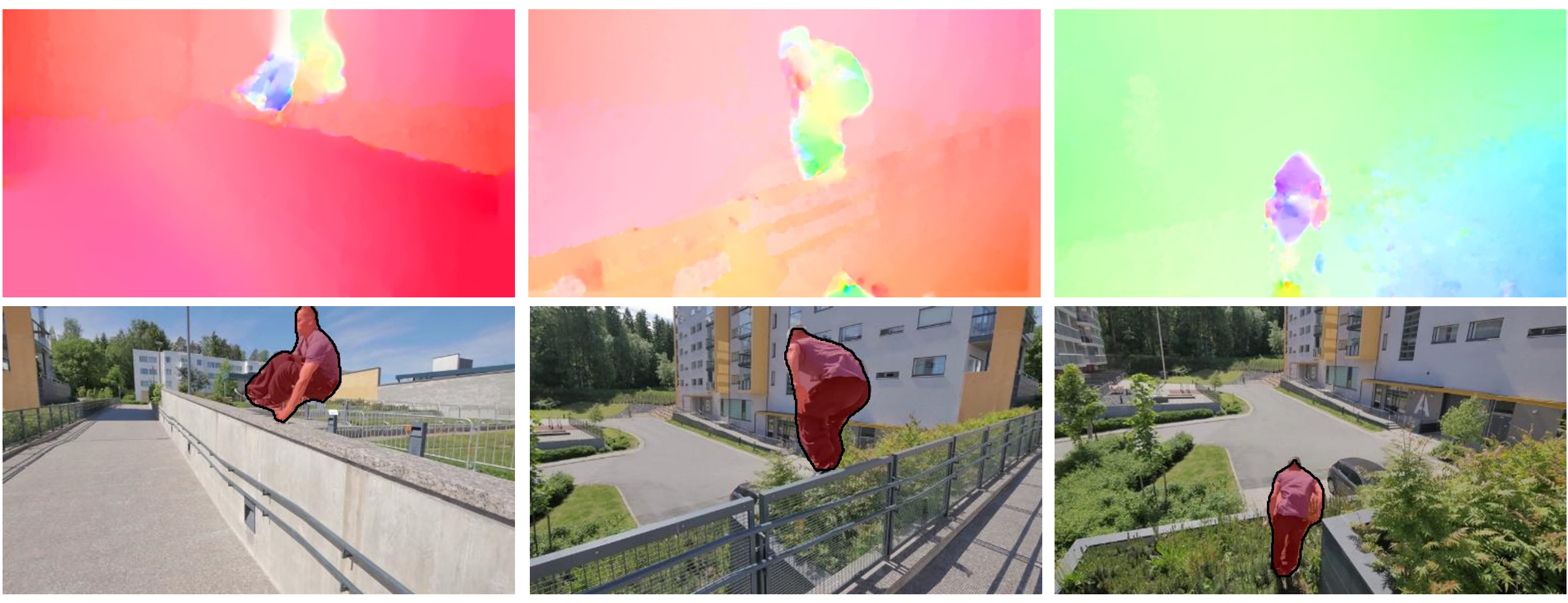}%change to network_2 for a single column version
\caption{We show color-coded optical flow images (first row) and video segmentation results (second row) produced by our joint model. Our proposed end-to-end trainable model simultaneously draws on the respective strengths of generic object appearance and motion in a unified framework.}
\label{fig:concept}
\vspace{-10pt}

\end{figure}

While appearance alone drives segmentation in images, videos provide a rich and complementary source of information in form of object motion.  It is natural to expect that both appearance and motion should play a key role in successfully segmenting objects in videos.   However, existing methods fall short of bringing these complementary sources of information together in a unified manner.

In particular, today motion is employed for video segmentation in two main ways.
On the one hand, the propagation or interactive techniques \emph{strongly rely on appearance} information stemming from human-drawn outlines on frames in the video. Here motion is primarily used to either propagate information or enforce temporal consistency in the resulting segmentation~\cite{sudheendra-eccv2012,suyog-eccv2014,Wen_2015_CVPR,Perazzi_2015_ICCV}. On the other hand, fully automatic methods \emph{strongly rely on motion} to seed the segmentation process by locating possible moving objects.  \KG{Once a moving object is detected, appearance is primarily used to track it across frames~\cite{keysegments,shah-cvpr2013,ferrari-iccv2013,nlc}.}  Such methods can fail if the object(s) are static or when there is significant camera motion.  In either paradigm, results suffer because the two essential cues are treated only in a sequential or disconnected way.

We propose an end-to-end trainable model that draws on the respective strengths of generic \KG{(non-category-specific)} object appearance and motion in a unified framework. %We are the first to show that genericcues about object appearance together with motion cues derived from optical flow can result in large gains in video object segmentation quality.
Specifically, we develop a novel two-stream fully convolutional deep segmentation network where individual streams encode generic appearance and motion cues derived from a video frame and its corresponding optical flow. These individual cues are fused in the network to produce a final object versus background pixel-level binary segmentation for each video frame.  
The proposed \CR{network} segments both static and moving objects \CR{in new videos} without any human involvement.

\KG{Declaring that motion should assist in video segmentation is non-controversial, and indeed we are certainly not the first to inject motion into video segmentation, as noted above.  However, % the empirical impact of motion for today's most challenging datasets has been arguably underwhelming. When it comes to motion and appearance,
thus far the sum is not much greater than its parts.  We contend that this is because \emph{the signal from motion is adequately complex such that rich learned models are necessary to exploit it}.  For example, a single object may display multiple motions simultaneously, background and camera motion can intermingle, and even small-magnitude motions should be informative.}

\KG{To learn the rich signals, sufficient training data is needed.  However, no large-scale video datasets with pixel-level segmentations exist.  Our second contribution is to address this practical issue.  We propose a solution that leverages readily available \emph{image} segmentation annotations together with \emph{weakly annotated video} data to train our model.}

Our results show the reward of learning from both signals in a unified framework: a true synergy, often with substantially stronger results than what we can obtain from either one alone---even if they are treated with an equally sophisticated deep network.  %show that indeed combining appearance and motion together in a unified framework results in a substantially better performance than what can be achieved individually. We
We \KG{significantly} advance the state-of-the-art for fully automatic video object segmentation on \KG{multiple} challenging datasets.  In some cases, the proposed method even outperforms existing methods that require manual intervention \CR{on the target video}. In summary our key contributions are:

\begin{itemize}
		\item \KG{the first} end-to-end \CR{trainable} framework for producing pixel level foreground object segmentation in videos.  	
		\item state-of-the-art on multiple datasets, improving over many reported results in the literature and strongly outperforming simpler applications of optical flow, and  
		\item a means to train a deep pixel-level video segmentation model with access to only weakly labeled videos and strongly labeled images, \KG{with no explicit assumptions about the categories present in either}.
% Our method shows that one can effectively leverage existing image segmentation datasets and combine them with weakly annotated videos to train an accurate model.
\end{itemize}

	\section{Related Work}

\paragraph{Automatic methods}
Fully automatic or unsupervised video segmentation methods assume no human input on the video. They can be grouped into two broad categories. First we have the supervoxel methods~\cite{grundmann-cvpr2010,xu-eccv2012,galasso-accv2012} which oversegment the video into space-time blobs with cohesive appearance and motion. Their goal is to generate mid-level video regions useful for downstream processing, whereas ours is to produce space-time tubes which accurately delineate object boundaries.  Second we have the fully automatic methods that generate \KG{thousands} of ``object-like" space-time segments~\cite{Wu_2015_CVPR,Fragkiadaki_2015_CVPR,Yu_2015_CVPR,oneata,xiao-cvpr2016}.
%. These include video region proposal methods which generate thousands of space-time segments per video \cc{either in form of bounding boxes or region tubes} using category independent properties of objects
While useful in accelerating object detection, it is not straightforward to automatically select the most accurate one when a single hypothesis is desired.  Methods that do produce a single hypothesis~\cite{keysegments,ma-cvpr2012,shah-cvpr2013,ferrari-iccv2013,nlc,Tsai_ECCV_2016,Sundberg,andrew_stein} strongly rely on motion to identify the objects, either by seeding appearance models with moving regions or directly reasoning about occlusion boundaries using optical flow. This limits their capability to segment static objects in video.  \KG{In comparison, our method is fully automatic, produces a single hypothesis, and can segment both static and moving objects.}  
%our method takes advantage of both generic object appearance and motion to identify the foreground objects and can successfully segment both static and moving objects.

\vspace*{-0.1in}
\paragraph{Human-guided methods}
Semi-supervised label propagation methods accept human input on a subset of frames, then propagate it to the remaining frames~\cite{ren-cvpr2007,tsai-bmvc2010,cipolla-cvpr2010,fathi-bmvc2011,sudheendra-eccv2012,suyog-eccv2014,Wen_2015_CVPR,Perazzi_2015_ICCV,marki2016bilateral,Tsai_CVPR_2016}. In a similar vein, interactive video segmentation methods leverage a human in the loop to provide guidance or correct errors, e.g.,~\cite{wang-tog2005,bai-2009,Nagaraja_2015_ICCV,price-iccv2009}. Since the human pinpoints the object of interest, these methods typically focus more on learning object appearance from the manual annotations. Motion is primarily used to propagate information or enforce temporal smoothness. In the proposed method, both motion and appearance play an equally important role, and we show their synergistic combination results in a much better segmentation quality. Moreover, our method is fully automatic and uses no human involvement \CR{to segment a novel video.}

%in segmenting the foreground object and we show that instead of relying on one or the other

\vspace*{-0.1in}
\paragraph{Category-specific semantic segmentation}
State-of-the-art semantic segmentation techniques for \emph{images} rely on fully convolutional deep learning architectures \cc{that are end-to-end trainable}~\cite{noh2015learning,crfasrnn_iccv2015,long_shelhamer_fcn,chen14semantic}. These deep learning based methods for segmenting images have seen rapid advances in recent years. Unfortunately, video segmentation has not seen such rapid progress. We hypothesize that the lack of large-scale human segmented video segmentation benchmarks is a key bottleneck.  Recent video benchmarks like Cityscapes~\cite{cityscapes} are valuable, but 1) it addresses category-specific segmentation, and 2) thus far methods competing on it process each frame independently, \cc{treating it like multiple image segmentation tasks.} In contrast, we aim to segment generic objects in video, whether or not they appear in training data.  Furthermore, our idea to leverage weakly labeled video for training opens a path towards training deep segmentation models that fuse spatial and temporal cues.
%that it is not optimal to treat video segmentation as an image segmentation problem. %%% NEITHER DO SOME METHODS...
%The synergy between motion and appearance when combined indeed results in a superior segmentation performance than what can be achieved with the individual parts.

\begin{figure*}[t]
\centering
\renewcommand{\tabcolsep}{0pt}
  \captionsetup{ font={footnotesize}, skip=2pt}
\includegraphics[width=2\columnwidth]{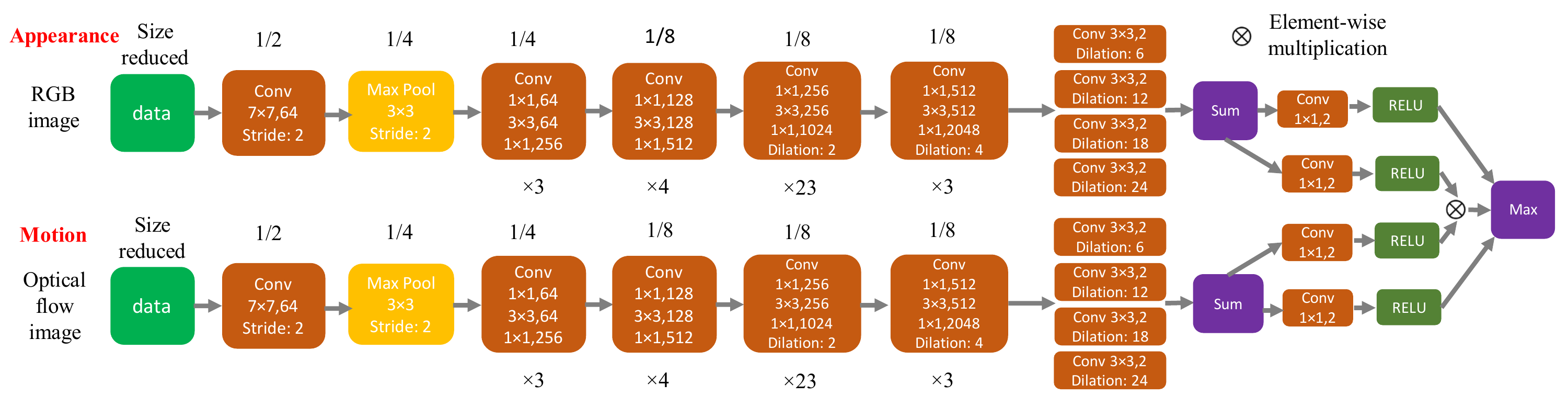}%change to network_2 for a single column version
\caption{Network structure for our model. Each convolutional layer except the first 7$\times$ 7  convolutional layer and our fusion blocks is a residual block~\cite{he2015deep}, adapted from ResNet-101. We show reduction in resolution at top of each box and the number of stacked convolutional layers in the bottom of each box.}
\label{fig:network}
\vspace{-10pt}
\end{figure*}

%\vspace*{-0.1in}
\paragraph{Deep learning with motion}
Deep learning for combining motion and appearance in videos has proven to be useful in several other computer vision tasks such as video classification~\cite{BeyondShort,KarpathyCVPR14}, action recognition~\cite{Simonyan,3DAction}, object tracking~\cite{DeepTrack,Wang_2015_ICCV,Ma-ICCV-2015} and even computation of optical flow~\cite{Dosovitskiy_2015_ICCV}. While we take inspiration from these works, we are the first to present a deep framework for segmenting objects in videos in a fully automatic manner.

%\CR{Some recent methods also use deep learning for semi-supervised video object segmentation~\cite{Valipour2016RecurrentFC,masktrack,oneshotvid,lucid}. All these techniques require human supervision on each video that needs to be segmented. In contrast, we are the first to present a unifying deep framework for segmenting objects in videos in a fully automatic manner.}

%\KG{CHECK OVER REFS-ARE MOST METHODS FROM RESULTS TABLES CITED IN HERE?  ENOUGH 2015 OR 2016 PAPERS THAT FIT THE ABOVE DESCRIPTIONS?}

\section{Approach}
\label{sec:approach}

Our goal is to segment generic objects in video, independent of the object categories they belong to, and without any manual intervention.   We pose the problem as a dense labeling task: given a sequence of video frames $[I_{1},I_{2},...,I_{N}]$, we want to infer either ``object" or ``background" for each pixel in each frame, to output a sequence of binary maps $[S_{1},S_{2},...,S_{N}]$.  %Posing video segmentation as a dense pixel labeling problem, 
We propose a solution based on a convolutional neural network.%% that utilizes both appearance and motion. %%\cc{Since large scale video datasets with pixel level segmentations are problematic, we show how to bootstrap weakly annotated videos together with strongly annotated image recognition datasets for training.}

First we segment generic objects based on appearance only from individual frames (Sec.~\ref{sec:appearance}). Then we use the appearance model to generate initial pixel-level annotations in training videos, and bootstrap strong annotations to train a model from motion (Sec.~\ref{sec:motion}). Finally, we fuse the two streams to perform video segmentation (Sec.~\ref{sec:joint}).

\subsection{Appearance Stream}\label{sec:appearance}

%While appearance provides valuable cues for generic video object segmentation, it 
%. However, it is very challenging to train such a model because the model should be able to predict a pixel-level map that aligns well with object boundaries and also generalize so it can assign high probability to pixels of unseen object categories.

%We follow the work of ~\cite{pixelObjectness} to train an appearance model that can take an image of arbitrary size and then produce a objectness map for each pixel. The key idea is to initialize the network using a powerful generic image representation learned from millions of images labeled by their object category and then fine-tune the network to produce dense binary object segmentation maps using relatively few images with pixel-level annotations. In particular, the deep fully convolutional neural network is initialized with weights pre-trained on ImageNet. Then we take a semantic segmentation dataset (PASCAL), and transform its dense semantic masks into binary object versus background masks, by fusing together all its object categories into a single supercategory(generic object). We then train the network to perform the dense foreground pixel labeling task.

Building on our ``pixel objectness" method~\cite{pixelObjectness}, we train a deep fully convolutional network to learn a model of \emph{generic foreground appearance}.  The main idea is to pre-train for object classification, then re-purpose the network to produce binary object segmentations by fine-tuning with relatively few pixel-labeled foreground masks.  Pixel objectness uses the VGG architecture~\cite{Simonyan14c} and transforms its fully connected layers into convolutional layers.  The resulting network possesses a strong notion of objectness, making it possible to identify foreground regions of more than 3,000 object categories despite seeing ground truth masks for only 20 during training.
%he main idea is to first train a network for object classification task with millions of images labeled by their object category. Then re-purpose the network to produce binary object segmentation by fine-tuning the network using relatively few images with pixel-level annotations. They argue that networks trained on large scale image classification task have already learned a strong notion of objectness, then by subsequently training with explicit dense foreground labels, the trained networks can produce high quality segmentation on generic objects. They also demonstrate their argument through experiments at a large scale.

%In particular, we initialize a deep fully convolutional neural network with weights pre-trained on ImageNet. Then we take a semantic segmentation dataset (PASCAL), and transform its dense semantic masks into binary object versus background masks, by fusing together all its object categories into a single supercategory(generic object). We then train the network to perform the dense generic object pixel labeling task.

%{\bf Model architecture:}
 %In ~\cite{pixelObjectness}, they adapt the widely used image classification model VGG-16 network~\cite{Simonyan14c} for structured prediction. 
 
 We take this basic idea and upgrade its implementation for our work.  In particular, we adapt the image classification model ResNet-101~\cite{he2015deep, chen2016deeplab} by replacing the last two groups of convolution layers with dilated convolution layers to increase feature resolution.  
 This results in only an 8$\times$ reduction in the output resolution instead of a 32$\times$ reduction in the output resolution in the original ResNet model.  In order to improve the model's ability to handle both large and small objects, \KG{we replace} the classification layer of ResNet-101 with four parallel dilated convolutional layers with different sampling rates to explicitly account for object scale. Then we fuse the prediction from all four parallel layers by summing all the outputs. The loss is the sum of cross-entropy terms over each pixel position in the output layer, where ground truth masks consist of only two labels---object foreground or background.  We train the model using the Caffe implementation of~\cite{chen2016deeplab}.  The network takes a video frame of arbitrary size and  produces an objectness map of the same size.  See Fig.~\ref{fig:network} (top stream).   
% \KG{ARE WE ADEQUATELY CITING OTHER WORK HERE BEYOND PIXEL OBJECTNESS REGARDING ARCHITECTURE?}
 
%\KG{While our method leverages pixel objectness~\cite{pixelObjectness} for the appearance stream, from this point forward in the paper our work cleanly departs from that contribution.  Whereas~\cite{pixelObjectness} tackles images, we tackle video.  The novelty of~\cite{pixelObjectness} is to learn an objectness model from few objects that generalizes to many others, with applications to image retrieval and retargeting.  The novelty of this paper is the first deep learning framework for \CR{fully automatic video object segmentation, and a means to train deep video segmentation models with weakly labeled video and strongly labeled images.}}

\subsection{Motion Stream}\label{sec:motion}

Our complete video segmentation architecture consists of a two-stream network in which parallel streams for appearance and motion process the RGB and optical flow images, respectively, then join in a fusion layer (see Fig.~\ref{fig:network}). 

The direct parallel to the appearance stream discussed above would entail training the motion stream to map optical flow maps to video frame foreground maps.  However, an important practical catch to that solution is training data availability.  While ground truth foreground image segmentations are at least modestly available, datasets for video object segmentation masks are small-scale in deep learning terms, and primarily support evaluation.  For example, Segtrack-v2~\cite{rehg-iccv2013}, a commonly used benchmark dataset for video segmentation, contains only 14 videos with 1066 labeled frames. DAVIS~\cite{Perazzi2016} contains only 50 sequences with 3455 labeled frames. None contain enough labeled frames to train a deep neural network. 
\KG{Semantic video segmentation datasets like CamVid~\cite{BrostowFC:PRL2008} or Cityscapes~\cite{cityscapes} are somewhat larger, yet limited in object diversity due to a focus on street scenes and vehicles.}  
%For semantic video segmentations, CamVid  contains 700 labeled images with 32 semantic classes and Cityscapes dataset contains 5000 images with fine annotations and 20,000 images with coarse annotations with 30 semantic classes. However, most semantic classes are restricted to constructions or vehicles. In addition, the two datasets are designed for semantic segmentation and have limited diversity for the task of video segmentation.} 
A good training source for our task would have ample frames with human-drawn segmentations on a wide variety of foreground objects, and would show a good mix of static and moving objects.  No such large-scale dataset exists and creating one is non-trivial.

We propose a solution that leverages readily available \emph{image} segmentation annotations together with \emph{weakly annotated video} data to train our model.  In brief, we temporarily decouple the two streams of our model, and allow the appearance stream to hypothesize likely foreground regions in frames of a large video dataset annotated only by bounding boxes.  Since appearance alone need not produce perfect segmentations, we devise a series of filtering stages to generate high quality estimates of the true foreground.  These instances bootstrap pre-training of the optical flow stream, then the two streams are joined to learn the best combination from minimal human labeled training videos.

%We observe that our trained appearance model performs exceedingly well on a large number of videos for segmenting generic objects. Our main idea is to apply our appearance model on video data to generate segmentations, which we then use as ground truth to train a motion model. Since our appearance might not produce perfect segmentation on all video data, we proposed to use a bounding box test to filter out unsuccessful segmentations to ensure ground truth quality.

More specifically, given a video dataset with bounding boxes labeled for each object,\footnote{We rely on ImageNet Video data, which contains 3862 videos and 30 diverse objects.  See Sec.~\ref{sec:results}.} we ignore the category labels and map the boxes alone to each frame.  Then, we apply the appearance stream, thus far trained only from images labeled by their foreground masks, to compute a binary segmentation for each frame.

\begin{figure}[t]
\centering

\renewcommand{\tabcolsep}{0pt}
  \captionsetup{ font={footnotesize}, skip=2pt}
\includegraphics[width=1\columnwidth]{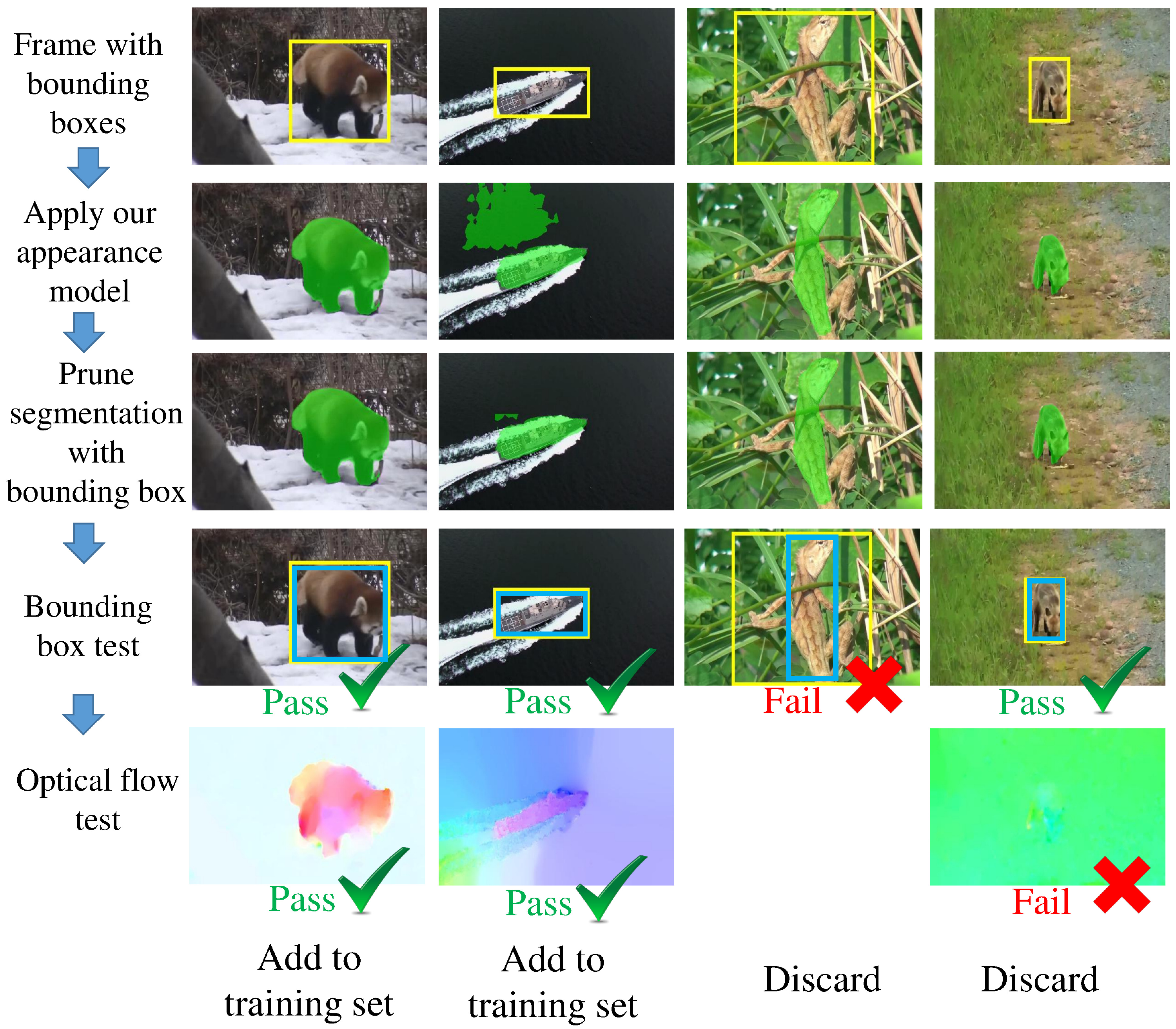}
\caption{Procedure to generate (pseudo)-ground truth segmentations. We first apply the appearance model to obtain initial segmentations (second row, with object segment in green) and then prune  by setting pixels outside bounding boxes as background (third row). Then we apply the bounding box test (fourth row, yellow bounding box is ground truth and blue bounding box is the smallest bounding box enclosing the foreground segment) and optical flow test (fifth row) to determine whether we add the segmentation to the motion stream's training set or discard it.  \KG{Best viewed in color.}}

\label{fig:step}
\vspace{-10pt}

\end{figure}

Next we deconflict the box and segmentation in each training frame.    
First, we refine the binary segmentation by setting all the pixels outside the bounding box(es) as background. 
Second, for each bounding box, we check if the the smallest rectangle that encloses all the foreground pixels overlaps with the bounding box by at least 75\%. Otherwise we discard the segmentation. 
Third, we discard regions where the box contains more than 95\% pixels labeled as foreground, based on the prior that good segmentations are rarely a rectangle, and thus probably the true foreground spills out beyond the box.  
Finally, we eliminate segments where object and background lack distinct optical flow,
so our motion model can learn from the desired cues.  Specifically, we compute the frame's optical flow using~\cite{liu2009beyond} and convert it to an RGB flow image~\cite{baker2011database}.  If the 2-norm between a) the average value within the bounding box and b) the average value in a box whose height and width are twice the original size exceeds 30, the frame and filtered segmentation are added to the training set. See Fig.~\ref{fig:step} for visual illustration of these steps.
%\footnote{threshold chosen by initial visual inspection}  

To recap, bootstrapping from the preliminary appearance model, followed by bounding box pruning, bounding box tests, and the optical flow test, we can generate accurate per-pixel foreground masks for thousands of diverse moving objects---for which no such datasets exist to date.  Note that by eliminating training samples with these filters, we aim to reduce label noise for training.  However, at test time our system will be evaluated on standard benchmarks for which each frame is manually annotated (see Sec.~\ref{sec:results}).

With this data, we now turn to training the motion stream.  
Analogous to our strong generic appearance model, we also want to train a strong generic motion model that can segment foreground objects purely based on motion. We use exactly the same network architecture as the appearance model (see Fig.~\ref{fig:network}). Our motion model takes only optical flow as the input and is trained with automatically generated pixel level ground truth segmentations. In particular, we convert the raw optical flow to a 3-channel (RGB) color-coded optical flow image~\cite{baker2011database}.  We use this color-coded optical flow image as the input to the motion network. We again initialize our network with pre-trained weights from ImageNet classification~\cite{ILSVRC15}.  Representing optical flow using RGB flow images allows us to leverage the strong pre-trained initializations as well as maintain symmetry in the appearance and motion arms of the network.

%\KG{ANYTHING ELSE TO SAY TO PRE-EMPT THE BIAS CONCERNS?  WE TRAIN WITH FG'S THAT COME FROM OUR OWN APPEARANCE STREAM, AND WE ARE FILTERING WHAT KINDS OF MOTIONS WE'LL EVEN GET TO LEARN FROM}
%\BX{Comments: There is only a small fraction of frames that are filtered out due to optical flow test. The motion test just makes sure that there is some motion in the optical flow image so the task of segmenting objects with motion is possible. In practical,the impact of optical flow test might be negligible. Optical flow test might not even be necessary. }
%\KG{We choose to first ``pre-train" the appearance and motion streams independently prior to fusion, in order to avoid squashing the effects of the motion stream.  SAY MORE ABOUT THIS...?}

An alternative solution might forgo handing the system optical flow, and instead input two raw consecutive RGB frames.  However, doing so would likely demand more training instances in order to discover the necessary cues.  Another alternative would directly train the joint model that combines both motion and appearance, whereas we first ``pre-train" each stream to make it discover convolutional features that rely on appearance or motion alone, followed by a fusion layer (below).  Our design choices are rooted in avoiding bias in training our model.  Since the (pseudo) ground truth comes from the initial appearance network, training jointly from the onset is liable to bias the network to exploit appearance at the expense of motion.   By feeding the motion model with only optical flow, we ensure our motion stream learns to segment objects from motion.

\subsection{Fusion Model}\label{sec:joint}

%We now describe how to fuse our appearance model and motion model. Note our motion model and appearance model have exactly the same network structure. We concatenate the two prediction outputs and then apply a 1$\times$1 convolution layer to obtain final prediction. See Fig. \ref{fig:network} for complete network structure. We do not fuse the two models in early stage of the networks because we want the two model to have independent predictions.

The final processing in our pipeline joins the outputs of the appearance and motion streams, and aims to leverage a whole that is greater than the sum of its parts.  We now describe how to train the joint model using both streams.

An object segmentation prediction is reliable if 1) either appearance or motion model alone predicts the object segmentation with very strong confidence or 2) their combination together predicts the segmentation with high confidence. This motivates the structure of our joint model. 

We implement the idea by creating three independent parallel branches: 1) We apply a 1$\times$1 convolution layer followed by a RELU to the output of the appearance model 2) We apply a 1$\times$1 convolution layer followed by a RELU to the output of the motion model 3) We replicate the structure of first and second branches and apply element-wise multiplication on their outputs. The element-wise multiplication ensures the third branch outputs confident predictions of object segmentation if and only if both  appearance model and motion model have strong predictions.   We finally apply a layer that takes the element-wise maximum to obtain the final prediction. See Fig. \ref{fig:network}.   

As discussed above, we do not fuse the two streams in an early stage because we want them both to have strong independent predictions.  Another advantage of our approach is we only introduce six additional parameters in each 1$\times$1 convolution layer, for a total of 24 trainable parameters. We can then train the fusion model with very limited annotated video data, without overfitting. \CR{In the absence of large volumes of video segmentation training data, precluding a complete end-to-end training, our strategy of decoupling the individual streams and training works very well in practice.}
	\section{Results}\label{sec:results}

\noindent {\bf Datasets and metrics:} We evaluate our method on three challenging video object segmentation datasets: DAVIS~\cite{Perazzi2016}, YouTube-Objects~\cite{prest2012learning,suyog-eccv2014,tang-cvpr2013} and Segtrack-v2~\cite{rehg-iccv2013}. To measure accuracy we use the standard Jaccard score, which computes the intersection over union overlap (IoU) between the predicted and ground truth object segmentations. The three datasets are:

\begin{itemize}
\vspace{-5pt}
\item {\bf DAVIS~\cite{Perazzi2016}:} the latest and most challenging video object segmentation benchmark consisting of 50 high quality video sequences of diverse object categories with $3,455$  densely annotated, pixel-accurate frames. The videos are unconstrained in nature and contain challenges such as occlusions, motion blur, and appearance changes. Only the prominent moving objects are annotated in the ground-truth. 
%While the videos contain both static and moving objects, only the prominent moving objects were annotated in the ground-truth. 
\vspace{-5pt}
\item {\bf YouTube-Objects~\cite{prest2012learning,suyog-eccv2014,tang-cvpr2013}:} consists of 126 challenging web videos from 10 object categories with more than 20,000 frames and is commonly used for evaluating video object segmentation. We use the subset defined in~\cite{tang-cvpr2013} and the ground truth provided by~\cite{suyog-eccv2014} for evaluation.
\vspace{-5pt}
\item {\bf SegTrack-v2~\cite{rehg-iccv2013}:} one of the most common benchmarks for video object segmentation consisting of 14 videos with a total of $1,066$ frames with pixel-level annotations.  For videos with multiple objects with individual ground-truth segmentations, we treat them as a single foreground for evaluation.

\end{itemize}

\begin{table*}[t]
	\centering
\footnotesize
	\begin{tabular}{|c|ccccc|ccc|ccc|}
		\hline
		\multicolumn{12}{|c|}{DAVIS: Densely Annotated Video Segmentation dataset (50 videos)}  \\
		\hline
		\hline
		%Methods & Flow-Th & Flow-Sal & FST & KEY & NLC & HVS & FCP & BVS & Ours-A & Ours-M & Ours-Joint \\
		Methods & Flow-Th & Flow-Sal & FST~\cite{ferrari-iccv2013} & KEY~\cite{keysegments} & NLC~\cite{nlc} & HVS~\cite{grundmann-cvpr2010} & FCP~\cite{Perazzi_2015_ICCV} & BVS~\cite{marki2016bilateral} & Ours-A & Ours-M & Ours-Joint \\
		
		\hline 
       \CR{Human in loop?} & No & No & No & No & No & Yes & Yes & Yes & No & No & No \\
		\hline
		\hline		
		Avg. IoU  & 42.95 & 30.22 & 57.5 & 56.9 & 64.1 & 59.6 & 63.1 & {\bf 66.5} & 64.69 & 60.18 & {\bf 71.51} \\
		\hline
	\end{tabular}
	\caption{Video object segmentation results on DAVIS dataset. We show the average accuracy over all 50 videos. Our method outperforms several state-of-the art methods, including the ones which actually require human annotations during segmentation. \CR{The best performing methods grouped by whether they require human-in-the-loop or not during segmentation are highlighted in bold.}  Metric: Jaccard score, higher is better. Please see supp. for per video results.}
	\label{davis-results}
\end{table*}

\noindent {\bf Baselines:} We compare with several state-of-the-art methods for each dataset as reported in the literature. Here we group them together based on whether they can operate in a fully automatic fashion (automatic) or require a human in the loop (semi-supervised) to do the segmentation:

\begin{itemize}
\vspace{-5pt}
\item {\bf Automatic methods:} Automatic video segmentation methods do not require any human involvement to segment new videos. Depending on the dataset, we compare with the following state of the art methods: FST~\cite{ferrari-iccv2013}, KEY~\cite{keysegments}, NLC~\cite{nlc} and COSEG~\cite{Tsai_ECCV_2016}. All use some form of unsupervised motion or objectness cues to identify foreground objects followed by post-processing to obtain space-time object segmentations.
	
%~\cite{keysegments,ma-cvpr2012,shah-cvpr2013,ferrari-iccv2013,nlc,Sundberg,andrew_stein}
\vspace{-5pt}
\item {\bf Semi-supervised methods:} Semi-supervised methods bring a human in the loop. They have some knowledge about the object of interest which is exploited to obtain the segmentation (e.g., a manually annotated first frame). We compare with the following state-of-the-art methods: HVS~\cite{grundmann-cvpr2010}, HBT~\cite{godec11a}, FCP~\cite{Perazzi_2015_ICCV}, IVID~\cite{Nagaraja_2015_ICCV}, HOP~\cite{suyog-eccv2014}, and BVS~\cite{marki2016bilateral}. The methods require different amounts of human annotation to operate, e.g.  HOP, BVS, and FCP make use of manual complete object segmentation in the first frame to seed the method; HBT requests a bounding box around the object of interest in the first frame; HVS, IVID require a human to constantly guide the algorithm whenever it fails. 
\end{itemize}

\noindent Note that our method requires human annotated data only during training. At test time it operates in a fully automatic fashion.  %and does not require human involvement on every video that it has to process. 
%In that context we place it somewhere in between the unsupervised and semi-supervised techniques in terms of the supervision required to obtain the segmentations. 
\KG{Thus, given a new video, we require equal effort as the automatic methods, and less effort than the semi-supervised methods.}

Apart from these comparisons, we also examine some natural baselines and variants of our method:

\begin{itemize}

\vspace{-5pt}
\item {\bf Flow-thresholding (Flow-Th):} To examine the effectiveness of motion alone in segmenting objects, we  adaptively threshold the optical flow in each frame using the flow magnitude.  Specifically, we compute the mean and standard deviation from the L2 norm of flow magnitude and use ``mean+unit std." as the threshold.
\vspace{-5pt}
\item {\bf Flow-saliency (Flow-Sal):} Optical flow magnitudes can have large variances, hence we also try a variant which normalizes the flow by applying a saliency detection method~\cite{jiang2013saliency} to the flow image itself. We use average thresholding to obtain the segmentation.
\vspace{-5pt}
\item {\bf Appearance model (Ours-A):} To quantify the role of appearance in segmenting objects, we obtain segmentations using only the appearance stream of our model. 
\vspace{-15pt}
\item {\bf Motion model (Ours-M):} To quantify the role of motion, we obtain segmentations using only the motion stream of our model. 
%Note that this stream only sees the optical flow image and has no information about the object's appearance.
\vspace{-5pt}
\item {\bf Joint model (Ours-Joint):} Our complete joint model that learns to combine both motion and appearance together to obtain the final object segmentation.

\end{itemize}

\begin{table*}[t]
	\centering
	\footnotesize
	\begin{tabular}{|c|cccc|ccc|ccc|}
		\hline
		\multicolumn{11}{|c|}{YouTube-Objects dataset (126 videos)}  \\
		\hline		
		\hline		
		%Methods & Flow-Th & Flow-Sal & FST & NLC & COSEG & HBT & PF-MRF & HOP & IVID & Ours-A & Ours-M & Ours-Joint \\
		Methods & Flow-Th & Flow-Sal & FST~\cite{ferrari-iccv2013} & COSEG~\cite{Tsai_ECCV_2016} & HBT~\cite{godec11a} &  HOP~\cite{suyog-eccv2014} & IVID~\cite{Nagaraja_2015_ICCV} & Ours-A & Ours-M & Ours-Joint \\
		
		\hline			
		\CR{Human in loop?} & No & No & No  & No  & Yes & Yes & Yes & No & No & No \\
		\hline		
		\hline		
		airplane (6) & 18.27 & 33.32 & 70.9 & 69.3 & 73.6 & 86.27 & {\bf 89} & {\bf 83.38} & 59.38 & 81.74 \\
		bird (6) & 31.63 & 33.74 & 70.6 & {\bf 76} & 56.1 & 81.04 & {\bf 81.6} & 60.89 & 64.06 & 63.84 \\
		boat (15) & 4.35 & 22.59 & 42.5 & 53.5 & 57.8 & 68.59 & {\bf 74.2} & {\bf 72.62} & 40.21 & 72.38 \\
		car (7) & 21.93 & 48.63 & 65.2 & 70.4 & 33.9 & 69.36 & {\bf 70.9} & 74.50 & 61.32 & {\bf 74.92} \\
		cat (16) & 19.9 & 32.33 & 52.1 & 66.8 & 30.5 & 58.89 & {\bf 67.7} & 67.99 & 49.16 & {\bf 68.43} \\
		cow (20) & 16.56 & 29.11 & 44.5 & 49 & 41.8 & 68.56 & {\bf 79.1} & {\bf 69.63} & 39.38 & 68.07 \\
		dog (27) & 17.8 & 25.43 & 65.3 & 47.5 & 36.8 & 61.78 & {\bf 70.3} & 69.10 & 54.79 & {\bf 69.48} \\
		horse (14) & 12.23 & 24.17 & 53.5 & 55.7 & 44.3 & 53.96 & {\bf 67.8} & {\bf 62.79} & 39.96 & 60.44 \\
		mbike (10) & 12.99 & 17.06 & 44.2 & 39.5 & 48.9 & 60.87 & {\bf 61.5} & 61.92 & 42.95 & {\bf 62.74} \\
		train (5) & 18.16 & 24.21 & 29.6 & 53.4 & 39.2 & 66.33 & {\bf 78.2} & {\bf 62.82} & 43.13 & 62.20 \\
		\hline
		Avg. IoU & 17.38 & 29.05 & 53.84 & 58.11 & 46.29 & 67.56 & {\bf 74.03} & {\bf 68.57} & 49.43 & 68.43 \\
		\hline
	\end{tabular}
	\caption{Video object segmentation results on YouTube-Objects dataset. We show the average performance for each of the 10 categories from the dataset. The final row shows an average over all the videos. Our method outperforms several state-of-the art methods, including the ones which actually require human annotation during segmentation. The best performing methods grouped by whether they require human-in-the-loop or not during segmentation are highlighted in bold. Metric: Jaccard score, higher is better. }
	%\CR{The best performing methods grouped by whether they require human-in-the-loop or not during segmentation are highlighted in bold.} Metric: Jaccard score, higher is better. }
	\label{youtube-results}
\end{table*}

\begin{table*}[t!]
	\centering
	\footnotesize
	\begin{tabular}{|c|ccccc|cc|ccc|}
		\hline
		\multicolumn{11}{|c|}{Segtrack-v2 dataset (14 videos)}  \\
		\hline		
		\hline		
		Methods & Flow-Th & Flow-Sal & FST~\cite{ferrari-iccv2013} & KEY~\cite{keysegments} & NLC~\cite{nlc} & HBT~\cite{godec11a} & HVS~\cite{grundmann-cvpr2010} & Ours-A & Ours-M & Ours-Joint \\
		
		\hline	
	    \CR{Human in loop?} & No & No & No & No & No & Yes & Yes & No & No & No \\
		\hline	
		Avg. IoU & 37.77 & 27.04 & 53.5 & 57.3 & {\bf 80\textsuperscript{*} } & 41.3 & {\bf 50.8} & 56.88 & 53.04 & 61.40 \\
		\hline	
	\end{tabular}
	\caption{Video object segmentation results on Segtrack-v2. We show the average accuracy over all 14 videos. Our method outperforms several state-of-the art methods, including the ones which actually require human annotation during segmentation. The best performing methods grouped by whether they require human-in-the-loop or not during segmentation are highlighted in bold. $^{*}$For NLC results are averaged over 12 videos as reported in their paper~\cite{nlc}. Metric: Jaccard score, higher is better. Please see supp. for per video results. } %\CR{The best performing methods grouped by whether they require human-in-the-loop or not during segmentation are highlighted in bold.} Metric: Jaccard score, higher is better. Please see supp. for per video results.}
	\label{segtrack-results}
	\vspace{-8pt}
\end{table*}

\begin{figure*}[t]
  \centering
  \footnotesize
  \renewcommand{\tabcolsep}{0pt}
  \captionsetup{width=1\textwidth, font={footnotesize}, skip=2pt}
   \begin{tabular}{cc}
    \multicolumn{2}{c}{\includegraphics[keepaspectratio=true,scale=0.093]{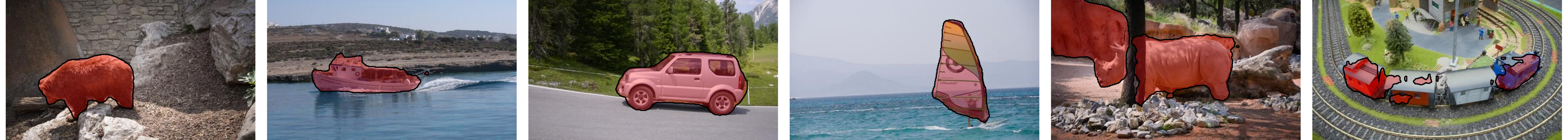}} \\
	\multicolumn{2}{c}{Appearance model (Ours-A)} \\    
    \multicolumn{2}{c}{\includegraphics[keepaspectratio=true,scale=0.093]{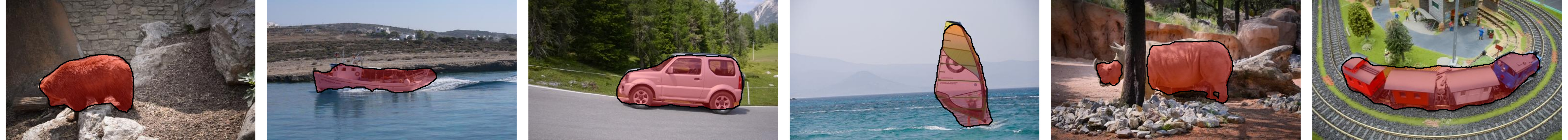}} \\    
	\multicolumn{2}{c}{Motion model (Ours-M)} \\    
    \multicolumn{2}{c}{\includegraphics[keepaspectratio=true,scale=0.093]{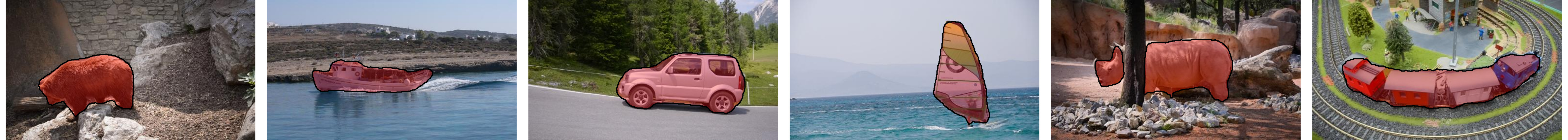}} \\
	\multicolumn{2}{c}{Joint model (Ours-Joint)} \\    
    \multicolumn{2}{c}{\includegraphics[keepaspectratio=true,scale=0.093]{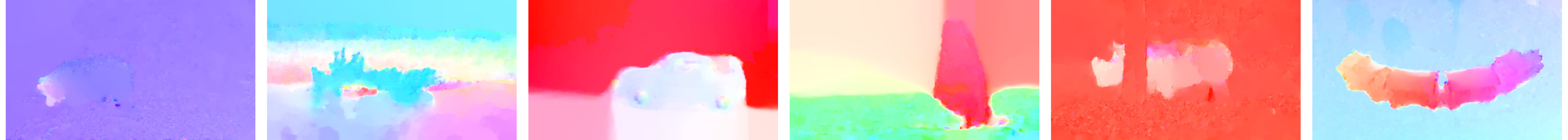}} \\    
	\multicolumn{2}{c}{Optical Flow Image} \\    
    \\
    {\small{\bf Ours vs. Automatic}} & {\small{\bf Ours vs. Semi-supervised}} \\
    \includegraphics[keepaspectratio=true,scale=0.093]{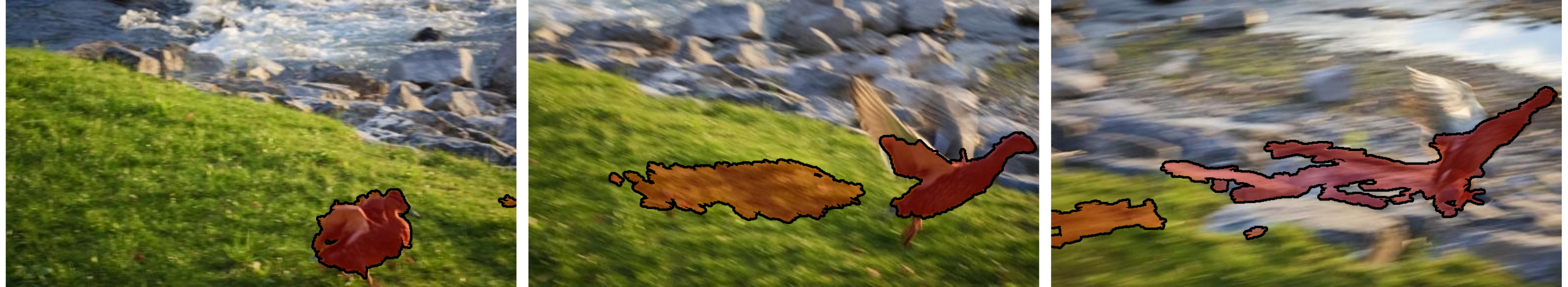} & \includegraphics[keepaspectratio=true,scale=0.093]{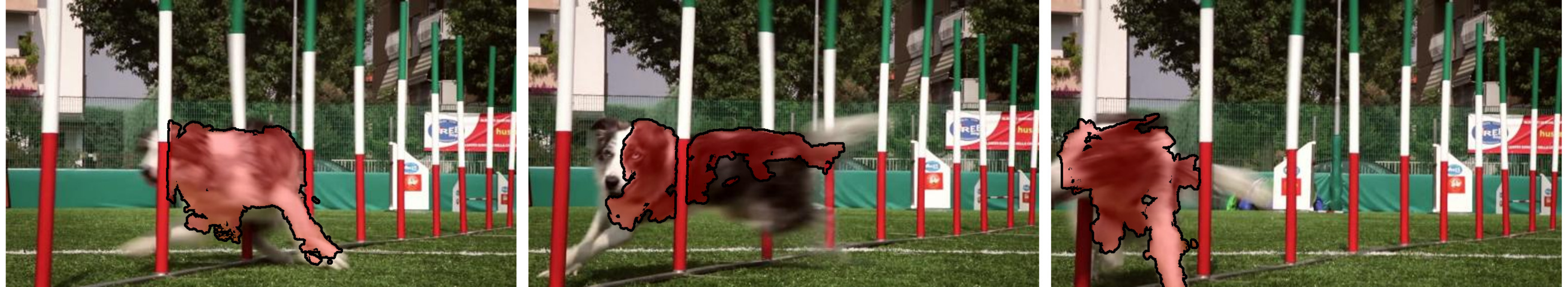} \\    
	 FST~\cite{ferrari-iccv2013} & BVS~\cite{marki2016bilateral} \\
    \includegraphics[keepaspectratio=true,scale=0.093]{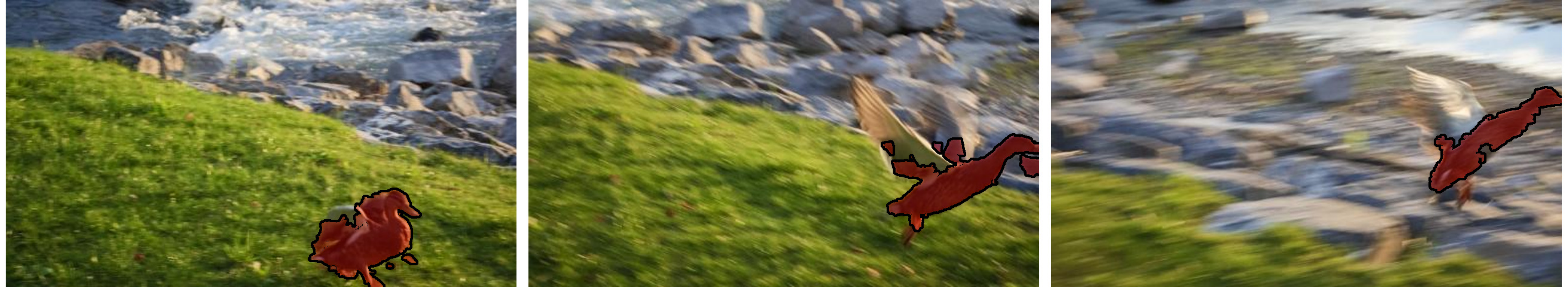} & \includegraphics[keepaspectratio=true,scale=0.093]{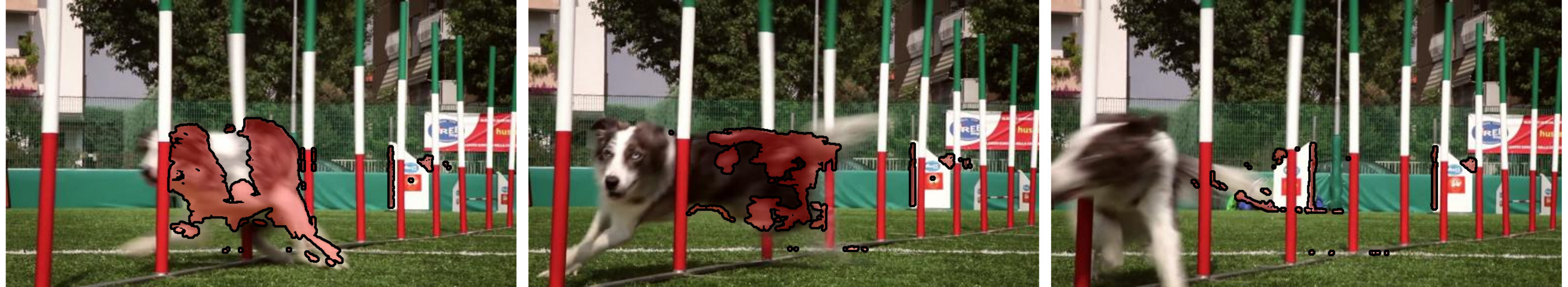} \\    
     NLC~\cite{nlc} & FCP~\cite{Perazzi_2015_ICCV} \\
    \includegraphics[keepaspectratio=true,scale=0.093]{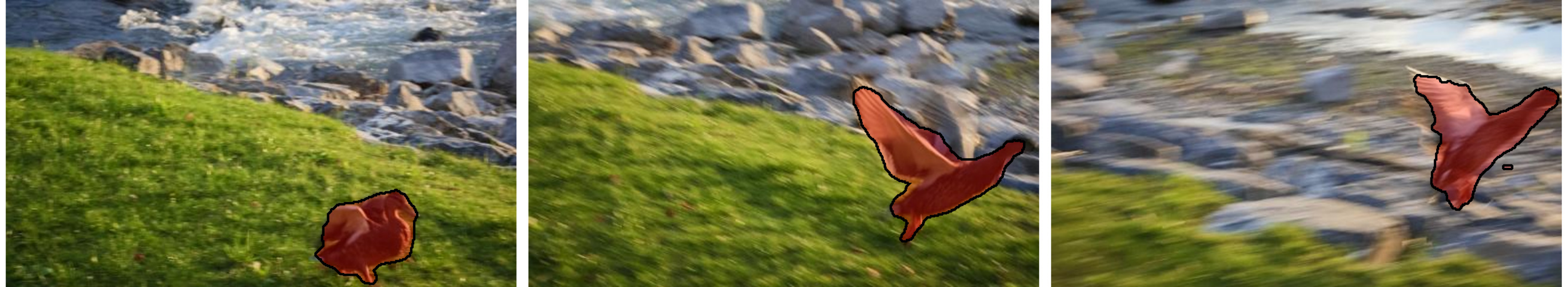} & \includegraphics[keepaspectratio=true,scale=0.093]{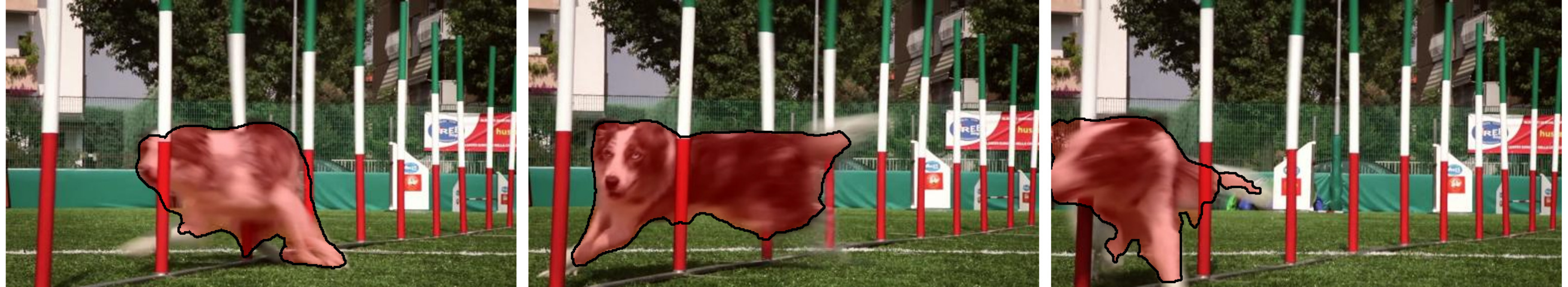} \\    
    Ours-Joint & Ours-Joint \\
	\end{tabular}
	\caption{Qualitative results: The top half shows examples from our appearance, motion, and joint models along with the flow image which was used as an input to the motion network.  The bottom rows show visual comparisons of our method with automatic and semi-supervised baselines (best viewed on pdf and see text for the discussion). Videos of our segmentation results are available on the project website. }
	\label{fig:qual_res}
\vspace{-5pt}
\end{figure*}

\noindent {\bf Implementation details:} To train the appearance stream, we rely on the PASCAL VOC 2012 segmentation dataset~\cite{pascal-seg} and use a total of 10,582 training images with binary object vs.~background masks (see~\cite{pixelObjectness} for more details). %As noted above, there is no large-scale video object segmentation dataset with pixel level ground truths available which can be directly used to train our motion stream.  Hence 
As weak bounding box video annotations, we use the ImageNet-Video dataset~\cite{ILSVRC15}.  This dataset comes with a total of 3,862 training videos from 30 object categories with 866,870 labeled object bounding boxes from over a million frames. Post refinement using our ground truth generation procedure (see Sec.~\ref{sec:motion}), we are left with 84,929 frames with good pixel segmentations\footnote{\CR{Available for download on our project website.}} which are then used to train our motion model. For training the joint model we use a held-out set for each dataset. \CR{We train each stream for a total of 20,000 iterations, use ``poly" learning rate policy (power = 0.9) with momentum (0.9) and weight decay (0.0005).  No post-processing is applied on the segmentations obtained from our networks. }

\vspace{5pt}
\noindent {\bf Quality of training data:} To ascertain that the quality of training data we automatically generate for training our motion stream is good, we first compare it with a small amount of human annotated ground truth.  We randomly select 100 frames that passed both the bounding box and optical flow tests, and collect human-drawn segmentations on Amazon MTurk.  We first present crowd workers a frame with a bounding box labeled for each object, and then ask them to draw the detailed segmentation for all objects within the bounding boxes. Each frame is labeled by three crowd workers and the final segmentation is obtained by majority vote on each pixel. %We use standard Jaccard score to evaluate the quality of our generated ground truth segmentation. 
The results indicate that our strategy to gather pseudo-ground truth is effective.  On the 100 labeled frames, Jaccard overlap with the human-drawn ground truth is 77.8 (and 70.2 before pruning with bounding boxes).  

%On the 100 labeled frames, direct prediction from our appearance model achieves a Jaccard score of 70.2.  After we prune the prediction with bounding boxes, the Jaccard score increase to 77.8. This demonstrate that our appearance model can produce good quality segmentation and our proposed bounding box pruning further improves the quality of segmentations.

\vspace{5pt}
\noindent {\bf Quantitative evaluation:} We now present the quantitative comparisons of our method with several state-of-the-art methods and baselines, for each of the three datasets in turn.

\vspace{5pt}
\noindent {\bf DAVIS dataset:} Table~\ref{davis-results} shows the results, with some of the best performing methods taken from the benchmark results~\cite{Perazzi2016}. Our method outperforms all existing methods on this dataset and significantly advances state-of-the-art. Our method is significantly better than simple flow baselines. This supports our claim that even though motion contains a strong signal about foreground objects in videos, it is not straightforward to simply threshold optical flow and obtain those segmentations. A data-driven approach that learns to identify motion patterns indicative of objects as opposed to backgrounds or camera motion is required. 

The appearance and motion variants of our method themselves result in a very good performance. The performance of the motion variant is particularly impressive, knowing that it has no information about object's appearance and purely relies on the flow signal. When combined together, the joint model results in a significant improvement, with an absolute gain of up to 11\% over individual streams.

Our method is also significantly better than fully automatic methods, which typically rely on motion alone to identify foreground objects. This illustrates the benefits of a unified combination of both motion and appearance.  Most surprisingly, our method significantly outperforms even the state-of-the-art semi-supervised techniques, which require substantial human annotation on every video they process. The main motivation behind bringing a human in the loop is to achieve higher accuracies than fully automated methods, yet in this case, our proposed fully automatic method outperforms the best human-in-the-loop algorithms by a significant margin. For example, the  BVS~\cite{marki2016bilateral} method---which is the current best performing semi-supervised method and requires the first frame of the video to be manually segmented---achieves an overlap score of 66.5\%. Our method significantly outperforms it with an overlap score of 71.51\%, yet uses no human involvement.% and thus significantly outperforms it. 

%\vspace*{0.1in}
\vspace{4pt}
\noindent {\bf  YouTube-Objects dataset:} In Table~\ref{youtube-results} we see a similarly strong result on the YouTube-Objects dataset. Our method again outperforms the flow baselines and all the automatic methods by a significant margin. The publicly available code for NLC~\cite{nlc} runs successfully only on 9\% of the YouTube dataset (1725 frames); on those, its jaccard score is 43.64\%. Our proposed model outperforms it by a significant margin of 25\%. Even among human-in-the-loop methods, we outperform all methods except IVID~\cite{Nagaraja_2015_ICCV}. However, IVID~\cite{Nagaraja_2015_ICCV} requires a human to consistently track the segmentation performance and correct whatever mistakes the algorithm makes. This can take up to minutes of annotation time for each video. Our method uses zero human involvement but still performs competitively.

It is also important to note that this dataset shares categories with the PASCAL segmentation benchmark which is used to train our appearance stream. Accordingly, we observe that the appearance stream itself results in the overall best performance. Moreover, this dataset has a mix of static and moving objects which explains the relatively weaker performance of our motion model alone. Overall the joint model works similarly well as appearance alone, however our ablation study (see Table~\ref{youtube_ablation}) where we rank test frames by their amount of motion, shows that our joint-model is stronger for moving objects. In short, our joint model outperforms our appearance model on moving objects, while our appearance model is sufficient for the most static frames. Whereas existing methods tend to suffer in one extreme or the other, our method handles both well.

\begin{table}[h!]
	\centering
	\footnotesize
	\begin{tabular}{|c|c|c|}
		\hline		
		Methods &  Top 10\% moving & Top 10\% static \\
		\hline	
		Ours-A  & 71.58 & 61.79 \\
		\hline	
		Ours-Joint  & 72.34 & 59.86 \\
		\hline	
	\end{tabular}
	\caption{Ablation study for YouTube-Objects dataset: Performance of our appearance and joint models on frames with most (left) and least (right) motion. }
	\label{youtube_ablation}
	\vspace{-8pt}
\end{table}

\vspace{5pt}
\noindent {\bf  Segtrack-v2 dataset:} In Table~\ref{segtrack-results}, our method outperforms all semi-supervised and automatic methods except NLC~\cite{nlc} on Segtrack. While our approach significantly outperforms NLC~\cite{nlc} on the DAVIS dataset, NLC is exceptionally strong on this dataset.  Our relatively weaker performance could be due to the low quality and resolution of the Segtrack-v2 videos, making it hard for our network based model to process them. Nonetheless, our joint model still provides a significant boost over both our appearance and motion models, showing it again realizes the synergy of motion and appearance in a serious way.
% benefiting from both appearance and motion.  
%\footnote{We put in our best effort to make the publicly available binary for NLC execute on YouTube-Objects but it fails and source code is not available. }

\vspace{5pt}
\noindent {\bf Qualitative evaluation:} Fig.~\ref{fig:qual_res} shows qualitative results. The top half shows visual comparisons between different components of our method including the appearance, motion, and joint models. We also show the optical flow image that was used as an input to the motion stream. These images help reveal the complexity of learned motion signals.  In the bear example, the flow is most salient only on the bear's head, still our motion stream alone is able to segment the bear completely. The boat, car, and sail examples show that even when the flow is noisy---including strong flow on the background---our motion model is able to learn about object shapes and successfully suppresses the background. The rhino and train examples show cases where the appearance model fails but when combined with the motion stream, the joint model produces accurate segmentations.  

The bottom half of Fig.~\ref{fig:qual_res} shows visual comparisons between our method and state-of-the-art automatic~\cite{ferrari-iccv2013,nlc} and semi-supervised~\cite{Perazzi_2015_ICCV,marki2016bilateral} methods. The automatic methods have a very weak notion about object's appearance; hence they completely miss parts of objects~\cite{nlc} or cannot disambiguate the objects from background~\cite{ferrari-iccv2013}. Semi-supervised methods~\cite{Perazzi_2015_ICCV,marki2016bilateral}, which rely heavily on the initial human-segmented frame to learn about object's appearance, start to fail as time elapses and the object's appearance changes considerably. In contrast, our method successfully learns to combine generic cues about object motion and appearance, segmenting much more accurately across all frames even in very challenging videos.

% FST~\cite{ferrari-iccv2013} 
% BVS~\cite{marki2016bilateral} 
% NLC~\cite{nlc} 
% FCP~\cite{Perazzi_2015_ICCV} 

	\section{Conclusions}
We presented a new approach for learning to segment generic objects in video that 1) achieves deeper synergy between motion and appearance and 2) addresses practical challenges in training a deep network for video segmentation.  Results show sizeable improvements over many existing methods---in some cases, even those requiring human intervention.  In future work we plan to explore extensions that could permit individuation of multiple touching foreground objects, as well as ways to incorporate human intervention intelligently into our framework. \\

\noindent {\bf Video examples, code \& pre-trained models available at:} \\ \url{http://vision.cs.utexas.edu/projects/fusionseg/} \\

\noindent {\bf Acknowledgements:} This research is supported in part by ONR YIP N00014-12-1-0754.
	\section{Appendix} 

\noindent {\bf Per-video results for DAVIS and Segtrack-v2:} Table~\ref{davis-results-supp} shows the per video results for the 50 videos from the DAVIS dataset. Table~\ref{davis-results} in the main paper summarizes these results over all 50 videos. We compare with several semi-supervised and fully automatic baselines. Our method outperforms the per-video best fully automatic and semi-supervised baseline in 25 out of 50 videos. 

Table~\ref{segtrack-results-supp} shows the per video results for the 14 videos from the Segtrack-v2 dataset. Table~\ref{segtrack-results} in the main paper summarizes these results over all 14 videos. Our method outperforms the per-video best fully automatic method in 5 out of 14 cases. Our method also outperforms the semi-supervised HVS~\cite{grundmann-cvpr2010} method in 8 out of 14 cases.

\begin{table*}[t]
	\centering
	\footnotesize
	 \captionsetup{ font={small}, skip=2pt}
	\begin{tabular}{|c|ccc|ccc|ccc|}
		\hline
		\multicolumn{10}{|c|}{DAVIS: Densely Annotated Video Segmentation dataset (50 videos)}  \\
		\hline
		\hline
		Methods & FST~\cite{ferrari-iccv2013} & KEY~\cite{keysegments} & NLC~\cite{nlc} & HVS~\cite{grundmann-cvpr2010} & FCP~\cite{Perazzi_2015_ICCV} & BVS~\cite{marki2016bilateral} & Ours-A & Ours-M & Ours-Joint \\
		
		\hline 
       Human in loop?  & No & No & No & Yes & Yes & Yes & No & No & No \\       
		\hline
		\hline		
		Bear & 89.8 & 89.1 & 90.7 & 93.8 & 90.6 & 95.5 & 91.52 & 86.30 & 90.66 \\
		Blackswan & 73.2 & 84.2 & 87.5 & 91.6 & 90.8 & 94.3 & 89.54 & 61.71 & 81.10 \\
		Bmx-Bumps & 24.1 & 30.9 & 63.5 & 42.8 & 30 & 43.4 & 38.77 & 26.42 & 32.97 \\
		Bmx-Trees & 18 & 19.3 & 21.2 & 17.9 & 24.8 & 38.2 & 34.67 & 37.08 & 43.54 \\
		Boat & 36.1 & 6.5 & 0.7 & 78.2 & 61.3 & 64.4 & 63.80 & 59.53 & 66.35 \\
		Breakdance & 46.7 & 54.9 & 67.3 & 55 & 56.7 & 50 & 14.22 & 61.80 & 51.10 \\
		Breakdance-Flare & 61.6 & 55.9 & 80.4 & 49.9 & 72.3 & 72.7 & 54.87 & 62.09 & 76.21 \\
		Bus & 82.5 & 78.5 & 62.9 & 80.9 & 83.2 & 86.3 & 80.38 & 77.70 & 82.70 \\
		Camel & 56.2 & 57.9 & 76.8 & 87.6 & 73.4 & 66.9 & 76.39 & 74.19 & 83.56 \\
		Car-Roundabout & 80.8 & 64 & 50.9 & 77.7 & 71.7 & 85.1 & 74.84 & 84.75 & 90.15 \\
		Car-Shadow & 69.8 & 58.9 & 64.5 & 69.9 & 72.3 & 57.8 & 88.38 & 81.03 & 89.61 \\
		Car-Turn & 85.1 & 80.6 & 83.3 & 81 & 72.4 & 84.4 & 90.67 & 83.92 & 90.23 \\
		Cows & 79.1 & 33.7 & 88.3 & 77.9 & 81.2 & 89.5 & 87.96 & 82.22 & 86.82 \\
		Dance-Jump & 59.8 & 74.8 & 71.8 & 68 & 52.2 & 74.5 & 10.32 & 64.22 & 61.16 \\
		Dance-Twirl & 45.3 & 38 & 34.7 & 31.8 & 47.1 & 49.2 & 46.23 & 55.39 & 70.42 \\
		Dog & 70.8 & 69.2 & 80.9 & 72.2 & 77.4 & 72.3 & 90.41 & 81.90 & 88.92 \\
		Dog-Agility & 28 & 13.2 & 65.2 & 45.7 & 45.3 & 34.5 & 68.94 & 67.88 & 73.36 \\
		Drift-Chicane & 66.7 & 18.8 & 32.4 & 33.1 & 45.7 & 3.3 & 46.13 & 44.14 & 59.86 \\
		Drift-Straight & 68.3 & 19.4 & 47.3 & 29.5 & 66.8 & 40.2 & 67.24 & 69.08 & 81.06 \\
		Drift-Turn & 53.3 & 25.5 & 15.4 & 27.6 & 60.6 & 29.9 & 85.09 & 72.09 & 86.30 \\
		Elephant & 82.4 & 67.5 & 51.8 & 74.2 & 65.5 & 85 & 86.18 & 77.51 & 84.35 \\
		Flamingo & 81.7 & 69.2 & 53.9 & 81.1 & 71.7 & 88.1 & 44.46 & 63.80 & 75.67 \\
		Goat & 55.4 & 70.5 & 1 & 58 & 67.7 & 66.1 & 84.11 & 74.99 & 83.09 \\
		Hike & 88.9 & 89.5 & 91.8 & 87.7 & 87.4 & 75.5 & 82.54 & 58.30 & 76.90 \\
		Hockey & 46.7 & 51.5 & 81 & 69.8 & 64.7 & 82.9 & 66.03 & 44.89 & 70.05 \\
		Horsejump-High & 57.8 & 37 & 83.4 & 76.5 & 67.6 & 80.1 & 71.09 & 54.10 & 64.93 \\
		Horsejump-Low & 52.6 & 63 & 65.1 & 55.1 & 60.7 & 60.1 & 70.23 & 55.20 & 71.20 \\
		Kite-Surf & 27.2 & 58.5 & 45.3 & 40.5 & 57.7 & 42.5 & 47.71 & 18.54 & 38.98 \\
		Kite-Walk & 64.9 & 19.7 & 81.3 & 76.5 & 68.2 & 87 & 52.65 & 39.35 & 49.00 \\
		Libby & 50.7 & 61.1 & 63.5 & 55.3 & 31.6 & 77.6 & 67.70 & 35.34 & 58.48 \\
		Lucia & 64.4 & 84.7 & 87.6 & 77.6 & 80.1 & 90.1 & 79.93 & 49.18 & 77.31 \\
		Mallard-Fly & 60.1 & 58.5 & 61.7 & 43.6 & 54.1 & 60.6 & 74.62 & 42.64 & 68.46 \\
		Mallard-Water & 8.7 & 78.5 & 76.1 & 70.4 & 68.7 & 90.7 & 83.34 & 25.31 & 79.43 \\
		Motocross-Bumps & 61.7 & 68.9 & 61.4 & 53.4 & 30.6 & 40.1 & 83.78 & 56.56 & 77.15 \\
		Motocross-Jump & 60.2 & 28.8 & 25.1 & 9.9 & 51.1 & 34.1 & 80.43 & 59.02 & 77.50 \\
		Motorbike & 55.9 & 57.2 & 71.4 & 68.7 & 71.3 & 56.3 & 28.67 & 45.71 & 41.15 \\
		Paragliding & 72.5 & 86.1 & 88 & 90.7 & 86.6 & 87.5 & 17.68 & 60.76 & 47.42 \\
		Paragliding-Launch & 50.6 & 55.9 & 62.8 & 53.7 & 57.1 & 64 & 58.88 & 50.34 & 57.00 \\
		Parkour & 45.8 & 41 & 90.1 & 24 & 32.2 & 75.6 & 79.39 & 58.51 & 75.81 \\
		Rhino & 77.6 & 67.5 & 68.2 & 81.2 & 79.4 & 78.2 & 77.56 & 83.03 & 87.52 \\
		Rollerblade & 31.8 & 51 & 81.4 & 46.1 & 45 & 58.8 & 63.27 & 57.73 & 69.01 \\
		Scooter-Black & 52.2 & 50.2 & 16.2 & 62.4 & 50.4 & 33.7 & 36.07 & 62.18 & 68.47 \\
		Scooter-Gray & 32.5 & 36.3 & 58.7 & 43.3 & 48.3 & 50.8 & 73.22 & 61.69 & 73.40 \\
		Soapbox & 41 & 75.7 & 63.4 & 68.4 & 44.9 & 78.9 & 49.70 & 53.24 & 62.57 \\
		Soccerball & 84.3 & 87.9 & 82.9 & 6.5 & 82 & 84.4 & 29.27 & 73.56 & 79.72 \\
		Stroller & 58 & 75.9 & 84.9 & 66.2 & 59.7 & 76.7 & 63.91 & 54.40 & 66.55 \\
		Surf & 47.5 & 89.3 & 77.5 & 75.9 & 84.3 & 49.2 & 88.78 & 73.00 & 88.41 \\
		Swing & 43.1 & 71 & 85.1 & 10.4 & 64.8 & 78.4 & 73.75 & 59.41 & 74.05 \\
		Tennis & 38.8 & 76.2 & 87.1 & 57.6 & 62.3 & 73.7 & 76.88 & 47.19 & 70.75 \\
		Train & 83.1 & 45 & 72.9 & 84.6 & 84.1 & 87.2 & 42.50 & 80.33 & 75.56 \\
		\hline
		\hline
		Avg. IoU  & 57.5 & 56.9 & 64.1 & 59.6 & 63.1 & {\bf 66.5} & 64.69 & 60.18 & {\bf 71.51} \\
		\hline
	\end{tabular}
	\caption{Video object segmentation results on DAVIS dataset. We show the results for all 50 videos. Table 1 in the main paper summarizes these results over all 50 videos. Our method outperforms several state-of-the art methods, including the ones which actually require human annotation during segmentation. The best performing methods grouped by whether they require human-in-the-loop or not during segmentation are highlighted in bold. Metric: Jaccard score, higher is better. }
	\label{davis-results-supp}
\end{table*}

\begin{table*}[t!]
	\centering
	\begin{tabular}{|c|ccc|c|ccc|}
		\hline
		\multicolumn{8}{|c|}{Segtrack-v2 dataset (14 videos)}  \\
		\hline		
		\hline		
		Methods  & FST~\cite{ferrari-iccv2013} & KEY~\cite{keysegments} & NLC~\cite{nlc}  & HVS~\cite{grundmann-cvpr2010} & Ours-A & Ours-M & Ours-Joint \\
		
		\hline	
		Human in loop?  & No & No & No & Yes  & No & No & No \\
		\hline	
		\hline
		birdfall2        & 17.50 & 49.00 & 74.00 & 57.40 & 6.94  & 55.50 & 38.01 \\
		bird of paradise & 81.83 & 92.20 & -     & 86.80 & 49.82 & 62.46 & 69.91 \\
		bmx              & 67.00 & 63.00 & 79.00 & 35.85 & 59.53 & 55.12 & 59.08 \\
		cheetah          & 28.00 & 28.10 & 69.00 & 21.60 & 71.15 & 36.00 & 59.59 \\
		drift            & 60.50 & 46.90 & 86.00 & 41.20 & 82.18 & 80.03 & 87.64 \\
		frog             & 54.13 & 0.00  & 83.00 & 67.10 & 54.86 & 52.88 & 57.03 \\
		girl             & 54.90 & 87.70 & 91.00 & 31.90 & 81.07 & 43.57 & 66.73 \\
		hummingbird      & 52.00 & 60.15 & 75.00 & 19.45 & 61.50 & 60.86 & 65.19 \\
		monkey           & 65.00 & 79.00 & 71.00 & 61.90 & 86.42 & 58.95 & 80.46 \\
		monkeydog        & 61.70 & 39.60 & 78.00 & 43.55 & 39.08 & 24.36 & 32.80 \\
		parachute        & 76.32 & 96.30 & 94.00 & 69.10 & 24.86 & 59.43 & 51.58 \\
		penguin          & 18.31 & 9.27  & -     & 74.45 & 66.20 & 45.09 & 71.25 \\
		soldier          & 39.77 & 66.60 & 83.00 & 66.50 & 83.70 & 48.37 & 69.82 \\
		worm             & 72.79 & 84.40 & 81.00 & 34.70 & 29.13 & 59.94 & 50.63 \\
		\hline 
		Avg. IoU  & 53.5 & 57.3 & {\bf 80\textsuperscript{*} }  & {\bf 50.8} & 56.88 & 53.04 & 61.40 \\
		\hline	
	\end{tabular}
	  \captionsetup{ font={small}, skip=2pt}
	\caption{Video object segmentation results on Segtrack-v2. We show the results for all 14 videos. Table 3 in the main paper summarizes these results over all 14 videos. Our method outperforms several state-of-the art methods, including the ones which actually require human annotation during segmentation. For NLC results are averaged over 12 videos as reported in their paper~\cite{nlc}. The best performing methods grouped by whether they require human-in-the-loop or not during segmentation are highlighted in bold.  Metric: Jaccard score, higher is better. }
	\label{segtrack-results-supp}
	\vspace{-8pt}
\end{table*}
	
	{\small
		\bibliographystyle{ieee}
		\bibliography{generic_object_extraction}
	}
	
	%this would normally be the end of your paper, but you may also have an appendix
	%within the given limit of number of pages
\end{document}